\pdfoutput=1

\documentclass[11pt]{article}

\usepackage{EMNLP2023}

\usepackage{times}
\usepackage{latexsym}

\usepackage[T1]{fontenc}

\usepackage[utf8]{inputenc}

\usepackage{microtype}

\usepackage{inconsolata}

\usepackage{graphicx}
\usepackage{subfigure}
\usepackage{bbding}

\usepackage{graphicx}               
\usepackage{tabularx}               
\usepackage{booktabs}
\usepackage{amsmath}
\usepackage{stmaryrd}
\usepackage{xcolor}
\usepackage{soul}

\newcolumntype{C}{>{\centering\arraybackslash}X}
\usepackage{multirow}               
\usepackage{diagbox}                
\usepackage{hhline}                 
\usepackage{color}                  
\usepackage{amsmath}                
\usepackage{mathtools}              
\usepackage{enumitem}               

\usepackage{subfigure}
\usepackage{booktabs}
\usepackage{extarrows}
\usepackage{makecell}

\usepackage{soul}

\definecolor{RoseQuartzBg}{HTML}{F7CAC9}
\definecolor{RoseQuartz}{HTML}{F5A798}
\definecolor{Serenity}{HTML}{92A8D1}
\definecolor{OrangeRed}{rgb}{1.0, 0.27, 0.0}
\definecolor{Red}{rgb}{1.0, 0.0, 0.0}
\definecolor{Turquoise}{HTML}{0F4C81}
\usepackage{xparse}
\NewDocumentCommand{\lifu}{ mO{} }{\textcolor{OrangeRed}{\textsuperscript{\textit{Lifu}}\textsf{\textbf{\small[#1]}}}}
\NewDocumentCommand{\minqian}{ mO{} }{\textcolor{violet}{\textsuperscript{\textit{Minqian}}\textsf{\textbf{\small[#1]}}}}
\NewDocumentCommand{\zhiyang}{ mO{} }{\textcolor{Serenity}{\textsuperscript{\textit{Zhiyang}}\textsf{\textbf{\small[#1]}}}}
\NewDocumentCommand{\ying}{ mO{} }{\textcolor{teal}{\textsuperscript{\textit{Ying}}\textsf{\textbf{\small[#1]}}}}

\NewDocumentCommand{\jy}{ mO{} }{\textcolor{brown}{\textsuperscript{\textit{jy}}\textsf{\textbf{\small[#1]}}}}

\usepackage[linesnumbered,ruled,vlined]{algorithm2e}
\usepackage{amsfonts}

\usepackage{graphicx}               
\usepackage{tabularx}               
\usepackage{booktabs}
\usepackage{amsmath}
\usepackage{stmaryrd}
\usepackage{xcolor}
\usepackage{soul}

\newcolumntype{C}{>{\centering\arraybackslash}X}
\usepackage{multirow}               
\usepackage{diagbox}                
\usepackage{hhline}                 
\usepackage{color}                  
\usepackage{amsmath}                
\usepackage{mathtools}              
\usepackage{enumitem} 

\newcommand{\jointEE}{\textsc{Joint4E-EE}}
\newcommand{\jointEL}{\textsc{Joint4E-EL}}

\newcommand{\vvv}[1]{\ensuremath{\mathbf{\lowercase{#1}}}} 
\newcommand{\mat}[1]{\ensuremath{\mathbf{\uppercase{#1}}}} 

\title{Iteratively Improving Biomedical Entity Linking and Event Extraction via Hard Expectation-Maximization}

\author{Xiaochu Li$^{*}$, Minqian Liu$^{*}$, \ Zhiyang Xu$^{*}$, \ Lifu Huang
\\
Computer Science Department \\
  Virginia Tech \\ 
  {\tt \{xiaocli, minqianliu, zhiyangx, lifuh\}@vt.edu}
  }

\begin{document}
\maketitle

\footnotetext[1]{Equal contribution.}

\begin{abstract}
Biomedical entity linking and event extraction are two crucial tasks to support text understanding and retrieval in the biomedical domain. These two tasks intrinsically benefit each other: entity linking disambiguates the biomedical concepts by referring to external knowledge bases and the domain knowledge further provides additional clues to understand and extract the biological processes, while event extraction identifies a key trigger and entities involved to describe each biological process which also captures the structural context to better disambiguate the biomedical entities. However, previous research typically solves these two tasks separately or in a pipeline, leading to error propagation. What's more, it's even more challenging to solve these two tasks together as there is no existing dataset that contains annotations for both tasks. 
To solve these challenges, we propose joint biomedical entity linking and event extraction by regarding the event structures and entity references in knowledge bases as latent variables and updating the two task-specific models in a hard Expectation–Maximization (EM) fashion: (1) predicting the missing variables for each partially annotated dataset based on the current two task-specific models, and (2) updating the parameters of each model on the corresponding pseudo completed dataset. Experimental results on two benchmark datasets: Genia 2011 for event extraction and BC4GO for entity linking, show that our joint framework significantly improves the model for each individual task and outperforms the strong baselines for both tasks. We will make the code and model checkpoints publicly available once the paper is accepted.
\end{abstract}

\section{Introduction} 


\begin{figure}[t]
\centering
\includegraphics[width=0.5\textwidth, trim={0.8cm 0 0 0},clip]{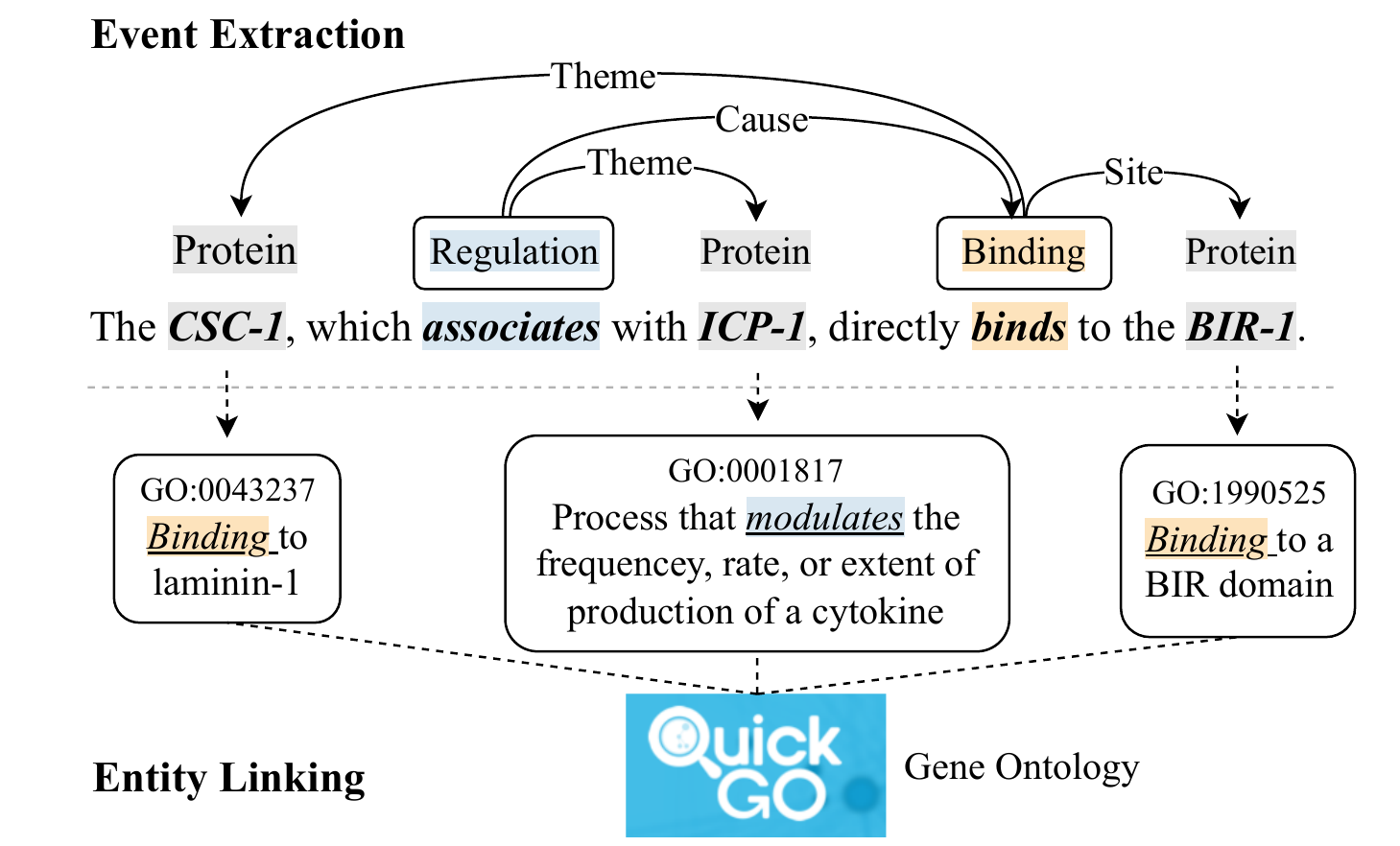}
\caption{
Illustration of biomedical entity linking (lower half) and event extraction (upper half) tasks given the same input. Below the input text, we show the definitions of each entity retrieved from Gene Ontology (GO) after running our entity linking model. We show the event types (in rounded boxes), entity types (without rounded boxes), and argument roles above the text. We highlight the event \textit{Regulation} and its mention in \textcolor{blue}{blue}, and the event \textit{Binding} and its mention in \textcolor{orange}{orange}. We also highlight the keywords in GO that are closely related to event extraction in corresponding colors.
}
\label{fig:Example}
\end{figure}

As the volume of biomedical literature continues to grow, biomedical entity linking and event extraction tasks have received increasingly more attention as they are essential to aid domain experts in retrieving and organizing critical information related to gene functions, bio-molecule relations, and bio-molecule behaviors from the vast amount of unstructured texts~\cite{kim2009overview,leitner2010overview,segura-bedmar-etal-2013-semeval}. Biomedical entity linking (a.k.a. named-entity disambiguation)~\cite{zhu2020latte,angell-etal-2021-clustering,dualencoder2021fast} aims to assign an entity mention in the text with a biomedical concept or term in reference biomedical knowledge bases, such as Gene Ontology (GO)~\cite{GOdata,BC4GOdata}, Unified Medical Language System (UMLS)~\cite{2004unified}, Universal Protein Resource (UniProt)~\cite{UniProdata}, and the EMBL nucleotide sequence database~\cite{EMBLdata}. Meanwhile, biomedical event extraction is the task of identifying \textit{event triggers} that most clearly convey the occurrence of events (i.e., biological processes) and their \textit{arguments} that participated in those events.
Figure~\ref{fig:Example} shows an example of biomedical entity linking and event extraction.  



Despite the recent progress achieved in biomedical entity linking and event extraction, there are still several problems that remained unaddressed. 
In biomedical entity linking, the entity mentions can be highly ambiguous as one mention can be mapped to multiple distinct biomedical concepts, requiring the model to have a good understanding of the context of the mention. For example, \textit{CSC-1} in Figure~\ref{fig:Example} can refer to a centromeric protein or a DNA \cite{2006chromosomal}. 
Meanwhile, biomedical events usually have complex and nested structures, and sufficient domain knowledge is required to capture biological processes and their participants. While each task has its own challenges, we find that these two tasks can be beneficial to each other:  entity linking maps the mentions in the text to biomedical concepts in external knowledge bases and provides additional domain knowledge and semantic information (e.g., the definitions in Gene Ontology) for extracting biological processes, while event extraction identifies the key trigger and its associated arguments that can provide more structural context to narrow down the pool of candidates and better link the entities to the biomedical concepts in knowledge bases. As shown in Figure~\ref{fig:Example}, the GO definition of the protein \textit{CSC-1} clearly indicates the function of this protein is related to the biological process \textit{binding}, which can help the event extraction model to infer the relationship between \textit{CSC-1} and the \textit{binding} event. On the other hand, given that \textit{CSC-1} is a \textit{Theme} of \textit{binding}, the entity linking can leverage such structural and precise context to better disambiguate the biological concept \textit{CSC-1}. 

While biomedical entity linking and event extraction intrinsically benefit each other, most existing works in biomedical information extraction ignore the close relationship between the two tasks and tackle them separately or in a pipeline, leading to the error propagation issue. 
Besides, there is no existing dataset that contains annotations for both tasks. For example, the BC4GO dataset~\cite{BC4GOdata} only contains the annotations for entity linking, whereas the Genia 11 dataset~\cite{genia11} only has the annotations for event extraction. This makes it even more difficult to solve these two tasks together.

To address these challenges, we propose a joint biomedical entity linking and event extraction framework, where each task-specific model incorporates the additional knowledge, i.e., the output from another model, to better perform task-specific prediction. 
To iteratively improve the models specific to each task, we model the entity references in knowledge bases and event structures as latent variables and devise a hard-EM-style learning strategy that consists of two steps: (1) \textbf{E-step:} estimating the missing variables for each partially annotated dataset (e.g., event triggers and their argument roles in the entity linking dataset) using the current two task-specific models, and; (2) \textbf{M-step:} updating the parameters of each model on the corresponding dataset that has been augmented by the pseudo labels in the complementary task. 

We extensively evaluate our approach on a biomedical entity linking dataset (i.e., BC4GO), and an event extraction dataset (i.e., Genia 11). The experimental results and case study validate the effectiveness of our approach. Our main contributions in this work are summarized as follows:
\begin{itemize}
    \item We propose a joint biomedical entity
linking and event extraction framework, namely \jointEL\  and \jointEE, where the two models can mutually improve each other.
    \item We design a collaborative training strategy to iteratively optimize two task-specific models such that each model can learn to leverage the information introduced by the other.
    \item Our joint framework consistently achieves significant performance gain for each individual task on four public benchmarks across different domains.
\end{itemize}
\section{Related Work}

\noindent\textbf{Biomedical Entity Linking.} 
Most recent state-of-the-art methods for biomedical entity linking are based on pre-trained BERT and consist of two steps: 
(1) candidate retrieval, which retrieves a small set of candidate references from a particular knowledge base; 
and (2) mention disambiguation and candidate ranking, which resolves the ambiguity of the mention based on the local context and refines the likelihood of each candidate reference with the fine-grained matching between mention and candidate~\cite{wu2019Blink,dualencoder2021fast,2021cross}. 
These methods are not efficient enough as it requires two pipelined models (a retrieval and a ranking model) and have shown to not be able to generalize well on rare entities~\cite{wu2019Blink,dualencoder2021fast}. 
Some recent studies have demonstrated that incorporating external information from biomedical knowledge bases, such as the latent type or semantic type information about mentions~\cite{zhu2020latte,xu-etal-2020-generate}, or infusing the domain-specific knowledge into the encoders with knowledge-aware pre-training tasks and objectives~\cite{2020infusing} can help improve the model performance on biomedical entity linking task~\cite{2021cross}. While these studies mainly leverage the knowledge from external knowledge bases to improve biomedical entity linking, related tasks such as biomedical event extraction can also provide meaningful clues to disambiguate the meaning of the mentions in the local context, however, it has not been previously studied, especially in the biomedical domain.


\noindent\textbf{Biomedical Event Extraction}
Current approaches for biomedical event extraction mainly focus on extracting triggers and arguments in a pipeline~\cite{eventbase2019,2019search,deepeventmine,sequencelabel,QandA2020}. Some studies also explore state-of-the-art neural methods with multiple classification layers to identify triggers, event types, arguments, and argument roles, respectively~\cite{eventbase2019,LSTM2019,graph2020biomedical,deepeventmine}.  
Recently, \cite{sequencelabel} propose a sequence labeling framework by converting the extraction of event structures into a sequence labeling task by taking advantage of a multi-label aware encoding strategy. In addition, to improve the generalizability of event extraction, \cite{QandA2020} establish a multi-turn question answering framework for event extraction by iteratively predicting answers for the template-based questions designed for event triggers, event arguments, and nested events. Several recent studies have also proposed to leverage external knowledge bases to disambiguate the biomedical terms in the local context and incorporate the knowledge, such as the definition or properties of the terms, into the event extraction process. Despite the success of these methods, they still suffer from error propagation in the pipeline frameworks, e.g., linking errors of biomedical terms in the local context will inform incorrect clues to the event extraction model and lead to a negative effect on the event predictions. Compared with all these studies, our \jointEE{} framework iteratively improves both biomedical entity linking and event extraction by leveraging the outputs from each other as additional input features.

\section{Problem Formulation}

\vspace{1mm}

\noindent\textbf{Biomedical Entity Linking.}
Given a text $\vvv{x}^L=[x_1^L,x_2^L,...,x_n^L]$ and a set of spans for all the entity mentions $\mathcal{M}=\{m_1,m_2,...,m_p\}$ in $\vvv{x}^L$, where $n$ indicates the number of tokens and $p$ indicates the number of mentions, biomedical entity linking maps each entity mention $m_i$ to a particular entity concept $\hat{c}_i$ from a biomedical knowledge base. Taking the sentence in Figure~\ref{fig:Example} as an example, for each entity mention, such as \textit{CSC-1}, a biomedical entity linking model will link it to a reference entity such as \textit{GO:0043237} in the external knowledge base of \textit{Gene Ontology}. Each entity in the knowledge base is represented with a unique GO ID and definition which is annotated by experts and Gene Ontology annotation tools\cite{gene2012gene,Go2013guide}. 

\vspace{1mm}

\noindent\textbf{Biomedical Event Extraction.}
Biomedical event extraction consists of two subtasks: event detection and argument extraction. Given the input text $\vvv{x}^E=[x_1^E,x_2^E,...,x_n^E]$, the goal of \textit{event detection} is to assign each token $x_i^E$ in $\vvv{x}^E$ with an event type $\tau_i$ that indicates a biological process in a predefined set of event types $\mathcal{T}$ or label it as \textit{Other} if the token is not an event trigger. 
For each identified event trigger, \textit{argument extraction} needs to assign each entity mention $m_i$ in $\mathcal{M}$ with an argument role $\alpha_j$ or \textit{Other} that indicates how the entity participates in the biological process $\tau_i$, where $\alpha_j$ belongs to a predefined set of argument role types $\mathcal{A}$. A mention is labeled as \textit{Other} if it does not participate in the particular biological processes triggered by $\tau_i$. As shown in Figure~\ref{fig:Example}, given the sentence as input, biomedical event extraction aims to detect all the candidate triggers and their types, such as \textit{associates} as a \textit{Regulation} event mention and \textit{binds} as a \textit{Binding} event mention, and extract the arguments with arguments roles for each trigger, e.g., \textit{ICP-1} is the \textit{Theme} of the \textit{associates} event while \textit{BIR-1} is the \textit{Site} of the \textit{binds} event. Note that, each event mention can also be an argument in another event, for example, \textit{associates} event is the \textit{Cause} of the \textit{binds} event. Thus, given a particular event trigger, we also predict an argument role $\alpha_j$ or \textit{Other} for each of the other triggers.


\section{Approach}

\begin{figure}[t]
\centering
\includegraphics[width=0.5\textwidth, trim={0 0 0 0},clip]{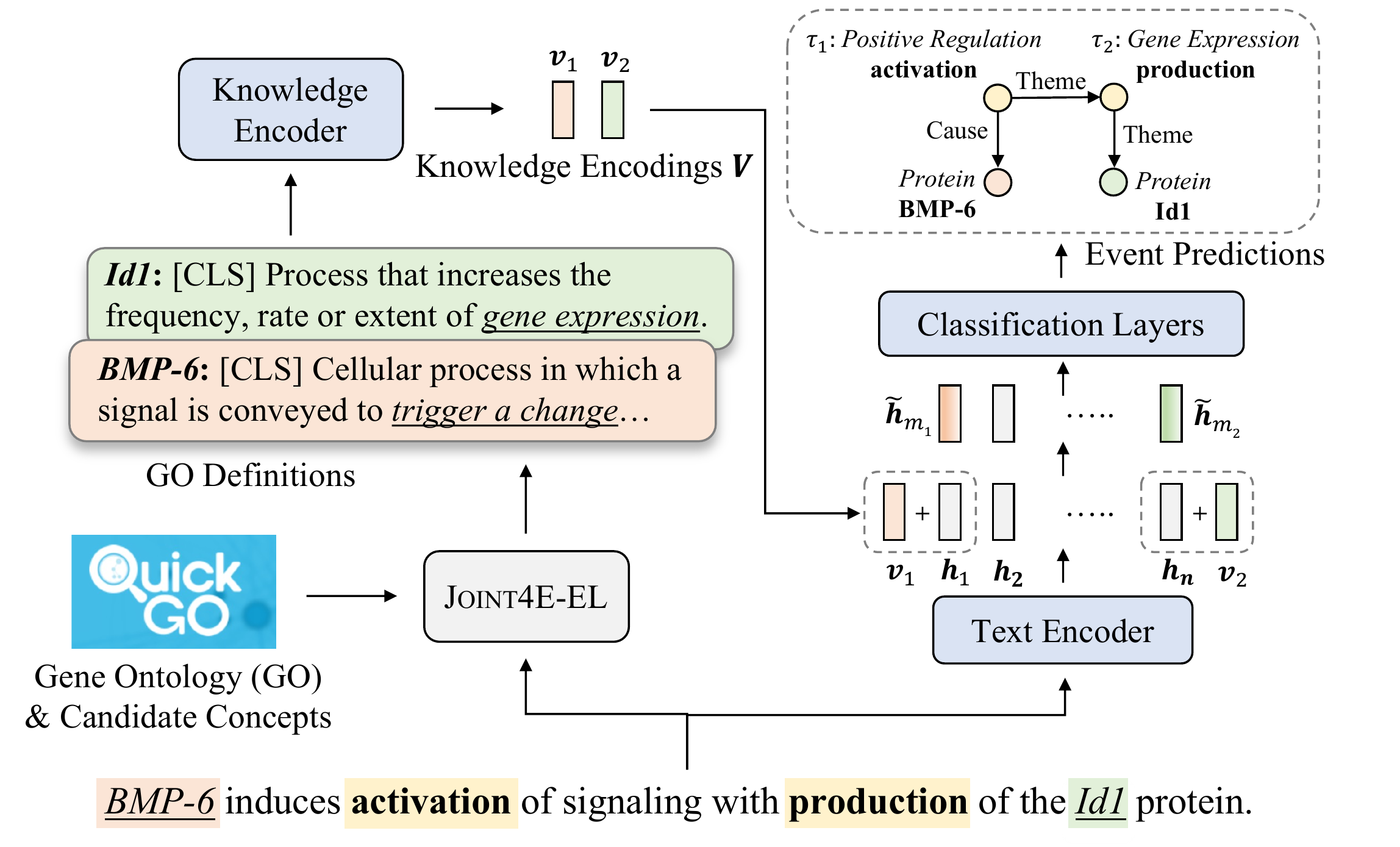}
\caption{
Illustration for our \jointEE{} for biomedical event extraction. \jointEE{} leverages the encoded GO definitions for each entity from the entity linking model \jointEL{} such that it has more domain knowledge to extract biological processes such as \textit{Gene Expression} and its participant \textit{Id1}.
}
\label{fig:jointee}
\end{figure}

In this section, we present our joint event extraction and entity linking framework that consists of (1) an entity-aware event extraction module, named \jointEE{}, that leverages the additional knowledge from knowledge bases, such as GO, UMLS, UniProt, and the EMBL nucleotide sequence database~\cite{GOdata,BC4GOdata,2004unified,UniProdata,EMBLdata}, to disambiguate the meaning of the biological terms in the input sentence so as to benefit the learning of the context and event extraction structures; and (2) an event-aware entity linking module, named \jointEL{}, which utilizes event structures to characterize the biological processes that each entity mention is involved and disambiguate its meaningful, so that we can better link each entity mention to the correct reference entity in the knowledge base. Since both \jointEL{} and \jointEE{} requires the output from the other task as additional input while there is no existing benchmark dataset containing annotations for both tasks, we further design a join training framework in an expectation-maximization (EM) fashion to iteratively estimate the missing variables (i.e., event structures or entity references from external knowledge base) and optimize both \jointEL{} and \jointEE{} simultaneously. In the following, we first introduce the details of \jointEL{} and \jointEE{} in Section~\ref{sec:jointEL} and~\ref{sec:jointEE}, and then elaborate on how we iteratively improve both task-specific models via an iterative learning schema in Section~\ref{sec:joint_training}.

\subsection{Entity-aware Biomedical Event Extraction (\jointEE{})} 
\label{sec:jointEE}
\noindent\textbf{Base Event Extraction Model.} The base event extraction model takes a text $\vvv{x}^E=[x_1^E,x_2^E,...,x_n^E]$ and the set of all entity mentions $\mathcal{M}$ in $\vvv{x}^E$ as inputs. We first encode $\vvv{x}^E$ with a PLM encoder \cite{bert,scibert} to obtain the contextualized representations $\mat{h}_w=[\vvv{h}_1,\vvv{h}_2,...,\vvv{h}_n]$ for the text, where each token's representation $\vvv{h}_j$ is the average of the representations of their corresponding subtokens. For each token $j$, we feed its representation $\vvv{h}_j$ into an event-type classification layer to classify the token into a positive event type or \textit{Other} if it is not an event trigger. Note that all event triggers are single-token. 

For argument extraction, we concatenate the contextualized representation of each identified event trigger $\vvv{h}_{\tau_j}$ with the representation of each argument candidate (i.e., entity mention) $\vvv{h}_{m_i}$ in $\mathcal{M}$ and feed the concatenated representations into an argument role classification layer to compute the probabilities for argument role types.
Both event detection and argument role classification are optimized with multi-class cross entropy.

\noindent\textbf{\jointEE{}.} For event extraction, we propose \jointEE{}, a dual encoder framework that incorporates the external domain knowledge of the given entities by the base entity linking model such that it can better extract biological processes from unstructured texts.
Given the input text $\vvv{x}^E$ and an entity mention $m_i$ from the set of all entity mentions $\mathcal{M}$, \textbf{first} we leverage the search engine of the QuickGO API\footnote{https://www.ebi.ac.uk/QuickGO/} for GO knowledge base to retrieve a set of candidate biomedical concepts $\mathcal{C}_i$ from the GO knowledge base. We type in the tokens of the entity mention to the search engine and the QuickGO API returns the set of all possible candidates. If there are more than 30 candidates returned, we only take the first 30 candidates returned by the QuickGO API. For the rest of the section, we use the term \textit{retrieve candidate concepts} to refer to the same process mentioned above. Table \ref{tab:fraction_candidate} shows the fraction of mentions that can be found with at least one positive candidate. When we take more than 30 candidates, the fraction doesn't increase. \textbf{Second}, we apply the base entity linking model to select a biomedical concept from the candidate set of concepts $\mathcal{C}_i$ retrieved from the GO knowledge base. \textbf{Third}, we obtain the definition of the corresponding biomedical concept from GO and use it as part of the input for \jointEE{}. In particular, we apply an additional PLM-based knowledge encoder that specifically takes in the selected biomedical definition $\vvv{d}_i=[d_{i,1},d_{i,2},...d_{i, q}]$ for $m_i$ and encodes it into contextualized representations. We take the contextualized representation of the [CLS] token as the knowledge encoding for $m_i$, denoted as $\vvv{v}_i$. We adopt the same process for all the entities in $\mathcal{M}$, which yields a set of knowledge encodings $\mathcal{V}=\{\vvv{v}_1, \vvv{v}_2,...,\vvv{v}_p\}$. Meanwhile, similarly to the base event extraction model, we also encode the input text $\vvv{x}^E$ into contextualized representations $\mat{h}_w=[\vvv{h}_1,\vvv{h}_2,...,\vvv{h}_n]$ with a text encoder. \textbf{Forth}, we integrate the external knowledge by applying element-wise addition between the contextualized representation of each mention $m_i$ and its corresponding knowledge encoding $\vvv{v}_i$ such that we obtain a knowledge-enhanced entity representation via $\tilde{\vvv{h}}_{m_i} = \vvv{h}_{m_i} + \vvv{v}_i$. \textbf{Finally}, we concatenate the representation of each identified event trigger $\vvv{h}_{\tau_j}$ with the enhanced entity representation $\tilde{\vvv{h}}_{m_i}$ and feed it into the classification layer to perform argument extraction.

\subsection{Event-aware Biomedical Entity Linking (\jointEL{})}
\label{sec:jointEL}
\noindent{\textbf{Base Entity Linking Model.}} The base entity linking model (Base-EL) takes in $\vvv{x}^L=[x_1^L,x_2^L,...,x_n^L]$ and the set of spans for all entity mentions $\mathcal{M}=\{m_1,m_2,...,m_p\}$ in $\vvv{x}^L$, and maps each entity mention $m_i\in\mathcal{M}$ to a concept in the external knowledge base, i.e., Gene Ontology (GO). We retrieve a set of candidate concepts $\mathcal{C}_i$ from GO for entity mention $m_i$. For each candidate $c_k$ from the candidate set $\mathcal{C}_i$, we obtain its definition in GO which is also a text sequence, denoted as $\vvv{d}_k=[d_{k,1},d_{k,2},...d_{k, q}]$. We append the definition $\vvv{d}_k$ at the end of $\vvv{x}^L$ separated by a special token [SEP], which yields the whole input sequence for the model:
\begin{equation}
    \text{[CLS]} [x_1^L,x_2^L,...,x_n^L]\text{[SEP]} [d_{k,1},d_{k,2},...d_{k, q}].
\end{equation}
We encode the entire sequence with a pretrained language model (PLM) encoder~\cite{bert, scibert} and then take the contextualized representation of the [CLS] token output from the encoder to compute the probability $$\mathbb{P}(c_k|m_i,\vvv{x}^L,\vvv{D}_k;\theta_{L})$$ with a binary classification layer, where $\theta_{L}$ denotes the parameters of the entity linking model. The model is optimized by the binary cross entropy loss. 
\textbf{\jointEL{}} We introduce \jointEL{}, a framework that utilizes the output of a base event extraction model (see Section 4.1 for details) for biomedical entity linking.
\jointEL{} consists of a PLM encoder \cite{bert,scibert} that computes the contextualized representations for the input sequence and a binary classification layer that computes the probability of the mapping between a given entity mention $m_i$ in the input text and a biomedical concept from a candidate set $\mathcal{C}_i$ in the knowledge base. 
Based on the base entity linking model, we incorporate the event information into the entity linking model to provide more structural context for better entity disambiguation. 

Specifically, given the input text $\vvv{x}^L$, \textbf{first}, we apply the base event extraction model to obtain the pseudo trigger and argument role labels. \textbf{Second}, we take one entity mention $m_i$ and retrieve candidate concepts $\mathcal{C}_i$ for $m_i$. \textbf{Third}, we inject event information into the input sequence $\vvv{x}^L$ of the entity linking model. Each entity only participates in a single event $\tau_i$ with a unique argument role $\alpha_{i}$ (if any). We insert the name of the argument role $\alpha_{i}$ after the tokens of $m_i$ in the original text $\vvv{x}^L$ and append the name of the event type $\tau_i$ at the end of the sentence $\vvv{x}^L$. Note that we set the name of the argument role as "\textit{Other}" if the entity does not participate in any biological process.
Similarly to the base entity linking model, we also append the definition $\vvv{d}_k$ w.r.t. the candidate concept $c_k$ after the original input. The event-enhanced input sequence for our \jointEL{} model is structured as: 
\begin{equation}
    \resizebox{0.95\hsize}{!}{ $\text{[CLS]} [x_1^L,x_2^L,...,m_i,\alpha_i,...,x_n^L, \tau_i]\text{[SEP]} [d_{k,1},d_{k,2},...d_{k, q}]$.}
\end{equation}
\textbf{Forth}, We encode the input sequence with the PLM encoder and feed the contextualized representation of the [CLS] token into the binary classification layer.

\begin{figure}[t]
\centering
\includegraphics[width=0.5\textwidth, trim={0 0 0 0},clip]{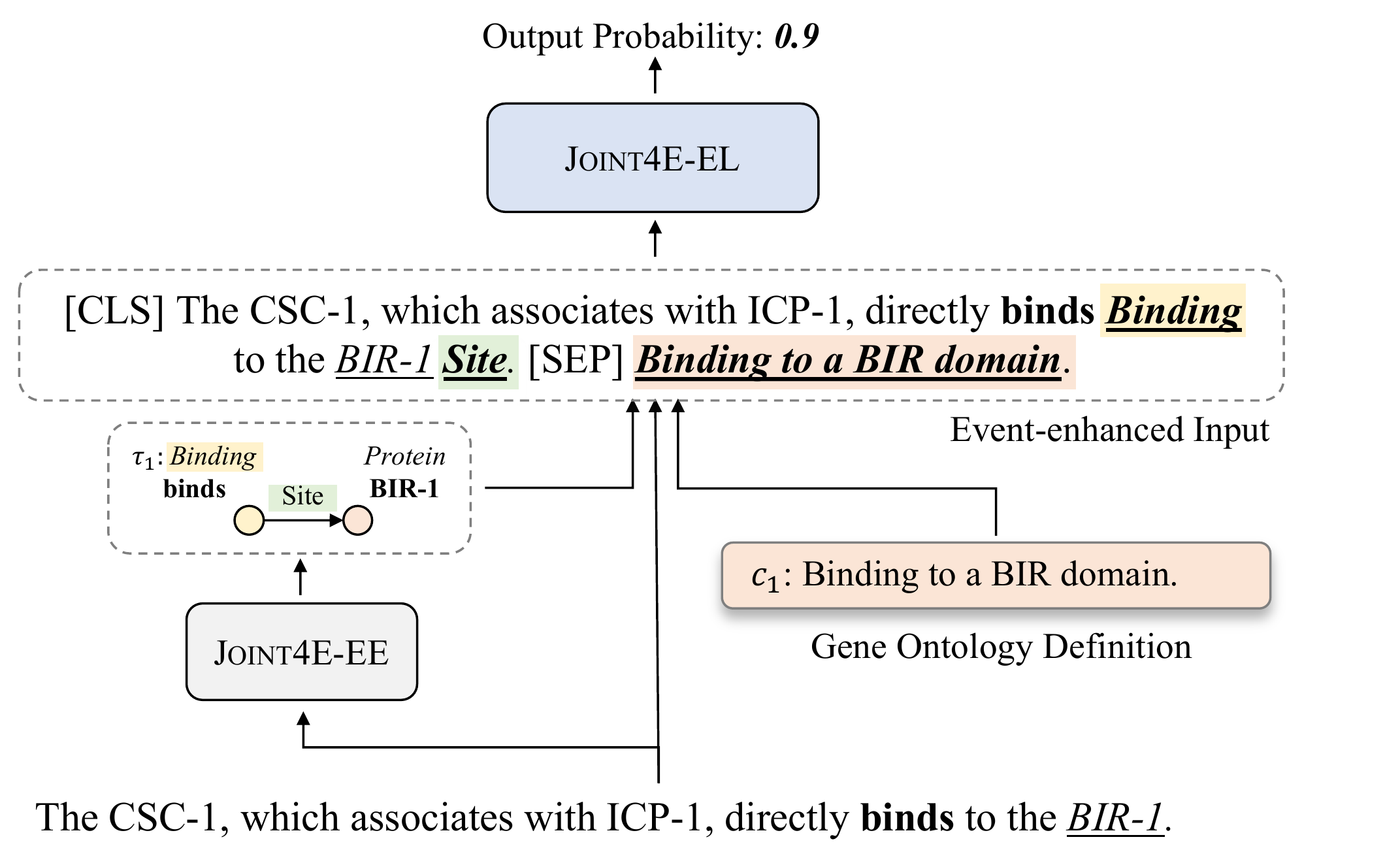}
\caption{
Illustration of our \jointEL{} for  biomedical entity linking. Given an entity mention (e.g., \textit{BIR-1}), \jointEL{} combines the original text, mention definition from Gene Ontology, and the predicted event structure from \jointEE{} as the event-enhanced input and outputs a probability to indicate its confidence on the candidate concept, e.g., $c_1$. We select the candidate with the highest probability as the predicted concept.
}
\label{fig:jointel}
\end{figure}
\begin{algorithm}[t]
\caption{Iterative Training for \textsc{Joint4E}}
\label{alg:iterative}

\KwIn{Entity linking dataset $\mathcal{D}_L$, event extraction dataset $\mathcal{D}_E$, external knowledge base $\mathcal{B}$, learning rates $\eta_L$ and $\eta_E$. }

\For{each entity set $\mathcal{M}$ in $\mathcal{D}_L$}{
    \For{$m_{i}\in\mathcal{M}$}{
        Retrieve the candidate set $\mathcal{C}_{i}$ for $m_{i}$ from $\mathcal{B}$\;
    }
}
Initialize the entity linking model's parameters $\theta_L$ and the event extraction model's  parameters $\theta_E$\;
Train $\theta_L$ on $\mathcal{D}_L$ and $\theta_E$ on $\mathcal{D}_E$\;

\While{not converged}{
    \tcp{Initialize augmented datasets}
    $\mathcal{U}_L=\{\}$, $\mathcal{U}_E=\{\}$ \;
    \tcp{E-step}
    \For{each $(\vvv{x}^L_i, \vvv{y}^L_i)\in\mathcal{D}_L$}{
        $\tilde{\vvv{z}}^E_i=\text{argmax}_{\vvv{z}_{j}^{E}\in \mathcal{Z}^E} \mathbb{P}(\vvv{z}_{j}^{E}|\vvv{x}^L_i;\theta_E)$\;
        $\mathcal{U}_L \leftarrow \mathcal{U}_L \bigcup \{(\vvv{x}^L_i, \vvv{y}^L_i, \tilde{\vvv{z}}^E_i)\}$\;
    }
    \For{each $(\vvv{x}^E_i, \vvv{y}^E_i)\in\mathcal{D}_E$}{
        $\tilde{\vvv{z}}^L_i=\text{argmax}_{\vvv{z}_{j}^{L}\in \mathcal{Z}^L} \mathbb{P}(z_{j}^{L}|x^E_i;\theta_L)$\;
        $\mathcal{U}_E \leftarrow \mathcal{U}_E \bigcup \{(\vvv{x}^E_i, \vvv{y}^E_i, \tilde{\vvv{z}}^L_i)\}$\;
    }
    \tcp{M-step}
    \For{each epoch}{
    Sample $(\vvv{x}^L_i, \vvv{y}^L_i, \tilde{\vvv{z}}^E_i)\sim\mathcal{U}_L$\;
    $\theta_L \leftarrow \theta_L - \eta_L\nabla_{\theta_L} J_L(\theta_L|\vvv{x}^L_i, \tilde{\vvv{z}}^E_i)$\;
    }
    \For{each epoch}{
    Sample $(\vvv{x}^E_i, \vvv{y}^E_i, \tilde{\vvv{z}}^L_i)\sim\mathcal{U}_E$\;
    $\theta_E \leftarrow \theta_E - \eta_E\nabla_{\theta_E} J_E(\theta_E|\vvv{x}^E_i, \tilde{\vvv{z}}^L_i)$\;
    }

}
\end{algorithm}

\subsection{Iterative Training via Hard-EM}
\label{sec:joint_training}
In this section, we formulate our Hard EM style iterative training algorithm which is shown in Algorithm~\ref{alg:iterative}.
For the event extraction task, we denote a training instance as $(\vvv{x}^E_i, \vvv{y}^E_i)$, where $\vvv{x}^E_i$ is a sentence and $\vvv{y}^E_i$ is the event annotation on  $\vvv{x}^E_i$, in the event extraction dataset $\mathcal{D}_E$. We denote $\mathcal{Z}_{all}^L$ as the finite set of all possible entity linking labels on the text $\vvv{x}^E_i$. We further define $\mathcal{Z}^L=\{\vvv{z}^L\in\mathcal{Z}_{all}^L:f_{\theta_E}(\vvv{x}^E_i,\vvv{z}^L)=\vvv{y}^E_i\}$ as the set of entity linking labels that leads to the correct event extraction prediction on the sentence $\vvv{x}^E_i$, where $f_{\theta_E}$ is the event extraction model and $\theta_E$ denotes its parameters. In our setting, the (pseudo) entity linking labels become discrete latent variables for the event extraction task.

For the entity linking task, given an instance $(\vvv{x}^L_i, \vvv{y}^L_i)$ where $\vvv{x}^L_i$ is a sentence and $\vvv{y}^L_i$ is the entity linking annotation on  $\vvv{x}^L_i$, in the entity linking dataset $\mathcal{D}_L$, we denote $\mathcal{Z}_{all}^E$ as the finite set of all possible event extraction labels on the text $\vvv{x}^L_i$. We further define $\mathcal{Z}^E=\{\vvv{z}^E\in\mathcal{Z}_{all}^E:f_{\theta_L}(\vvv{x}^L_i,\vvv{z}^E)=\vvv{y}^L_i\}$ as the set of event extraction labels that leads to the correct entity linking predictions on the sentence $\vvv{x}^L_i$, where $f_{\theta_L}$ is the entity linking model and $\theta_L$ denotes its parameters. In the above setting, the (pseudo) event extraction labels become discrete latent variables for the entity linking task.

Given a dataset $\mathcal{D}_L$ with entity linking annotations and a dataset $\mathcal{D}_E$ with event extraction annotations, we first perform the following prerequisite steps: \textbf{First}, We prepare the candidate biomedical concepts for both entity linking dataset $\mathcal{D}_L$ and event extraction dataset $\mathcal{D}_E$. \textbf{Second}, we randomly initialize the parameter $\theta_L$ for \jointEE{} and the parameter $\theta_E$ for \jointEL{}. To first obtain a well-initialized base model for each task, we individually train \jointEL{} on the labeled entity linking dataset $\mathcal{D}_L$ and train \jointEE{} on the labeled entity linking dataset $\mathcal{D}_L$ until the model converges on the development sets, respectively.
After we obtain a base model individually trained on each task, 
we start our Hard-EM style iterative training process that repeatedly performs the following two steps: (1) the E-step that aims to estimate the latent variables (i.e., predict pseudo labels) for each partially annotated dataset, and; (2) M-step where it updates the parameters of each model given the original inputs and the estimated latent variables.

\textbf{E-Step.} At the beginning of each round of the iterative training, we first initialize two empty sets $\mathcal{U}_L=\{\}$ and $\mathcal{U}_E=\{\}$ for collecting pseudo labeled instances. We run the entity linking model \jointEL{} on the event extraction dataset $\mathcal{D}_E$ to generate pseudo entity linking annotations. Specifically, for each instance in the event extraction dataset $(\vvv{x}^E_i, \vvv{y}^E_i)\in\mathcal{D}_E$, we run the \jointEL{} model and predict the pseudo entity linking labels $\mathcal{Z}^L$. Following hard EM, for the event extraction task, we take the latent variable $\Tilde{\vvv{z}}^L_i\in\mathcal{Z}^L$ that has the highest likelihood, i.e., $\tilde{\vvv{z}}^L_i=\text{argmax}_{\vvv{z}_{j}^{L}\in \mathcal{Z}^L} \mathbb{P}(z_{j}^{L}|x^E_i;\theta_L)$. The estimated latent variable $\tilde{\vvv{z}}^L_i$ together with $\vvv{x}^E_i$ and $\vvv{y}^E_i$ form a new instance and is added into $\mathcal{U}_E$.
We also run the event extraction model \jointEE{} on the entity linking dataset $\mathcal{D}_E$ to generate pseudo event extraction annotations. Specifically, for each instance in the entity linking dataset $(\vvv{x}^L_i, \vvv{y}^L_i)\in\mathcal{D}_L$, we run the \jointEE{} model and predict the pseudo event labels $\mathcal{Z}^E$. Following hard EM, for the event extraction task, we take the latent variable $\Tilde{\vvv{z}}^E_i\in\mathcal{Z}^E$ that has the highest likelihood, i.e., $\tilde{\vvv{z}}^E_i=\text{argmax}_{\vvv{z}_{j}^{E}\in \mathcal{Z}^E} \mathbb{P}(z_{j}^{E}|x^E_i;\theta_E)$. The estimated latent variable $\tilde{\vvv{z}}^E_i$ together with $\vvv{x}^L_i$ and $\vvv{y}^L_i$ form a new instance and is added into $\mathcal{U}_L$.

\textbf{M-Step.} For the event extraction task, we loop through the examples $(\vvv{x}^E_i, \vvv{y}^E_i, \tilde{\vvv{z}}^L_i)$ in the newly collected $\mathcal{U}_E$ event extraction dataset enhanced with pseudo entity linking annotations. The \jointEE{} model $f_\theta^E$ optimizes the log-likelihood of the true event extraction label $\vvv{y}^E_i$ based on the discrete latent variable $\tilde{\vvv{z}}^L_i$, i.e., the entity linking pseudo label. The loss is computed as $J_E(\theta_E|\vvv{x}^E_i, \tilde{\vvv{z}}^L_i)=-\log{\mathbb{P}(\vvv{y}^E_i|\vvv{x}^E_i,\tilde{\vvv{z}}^L_i;\theta_E)}$.
For the entity linking task, we loop through the examples $(\vvv{x}^L_i, \vvv{y}^L_i, \tilde{\vvv{z}}^E_i)$ in the newly collected $\mathcal{U}_L$ entity linking dataset enhanced with pseudo event extraction annotations. The \jointEL{} model $f_\theta^L$ optimizes the log-likelihood of the true entity linking label $\vvv{y}^L_i$ based on the discrete latent variable $\tilde{\vvv{z}}^E_i$, i.e., the event pseudo label. The loss is computed as $J_L(\theta_L|\vvv{x}^L_i, \tilde{\vvv{z}}^E_i)=-\log{\mathbb{P}(\vvv{y}^L_i|\vvv{x}^L_i,\tilde{\vvv{z}}^E_i;\theta_L)}$.\\

\section{Experimental setup}
\subsection{Datasets}

\paragraph{Event Extraction}
We evaluate the performance of our approach for biomedical event extraction on the Genia 2011 dataset (GE11)~\cite{genia11}, which defines 9 event types with 6 argument roles. The text is based on the abstracts and full articles from PubMed about biological processes related to proteins and genes.  The detailed statistics of GE11 are summarized in Table~\ref{tab:ge11data}. Following previous studies~\cite{sequencelabel,deepeventmine, QandA2020, graph2020biomedical,Zhao2021hann,wang2022cpje}, we evaluate the performance of biomedical event extraction using the precision (P), recall (R), and F1 score (F1). 


\begin{table}[t]
\begin{center}
\begin{tabular}{ l | c | c | c  }
\toprule
 GE11 & Training & Development & Test \\
\midrule
\# Documents & 908 & 259 & 347 \\
\# Sentences & 8,664 & 2,888 & 3,363 \\
\# Entities & 11,625 & 4,690 & 5,301 \\
\# Events & 10,310 & 3,250 & 4,487 \\
\bottomrule
\end{tabular}
\end{center}
\caption{Statistics of the Genia 2011 dataset for biomedical event extraction.}
\label{tab:ge11data}
\end{table}

\paragraph{Entity Linking}
For the entity linking task, we leverage the BioCreative IV GO (BC4GO) dataset \cite{BC4GOdata} which contains annotations of Gene Ontology entities for all the entity mentions in the dataset. Each entity mention in BC4GO is mapped to a unique biomedical entity in the Gene Ontology knowledge base where each entity is described with GO id, name, and definition. 
However, the original BC4GO dataset was built in 2013. With the development of Vivo and Vitro in biomedical science in the last decades, new definitions and ontologies of biomedical concepts have been introduced into the Gene Ontology knowledge base, which drastically changes the topology of the knowledge base~\cite{park2011gochase,yon2008use} and makes the mappings between the entity mentions and their concepts in the original BC4GO outdated. 
In addition, previous studies~\cite{Go2013guide} also suggest that the mappings between entity mentions and entities in the Gene Ontology knowledge base are not surjective, i.e., each entity mention can be mapped into multiple entities. 
Thus, we propose to update the mappings between entity mentions in BC4GO and entities in Gene Ontology by leveraging the official API~\footnote{\url{https://www.ebi.ac.uk/QuickGO/}} of Gene Ontology. We include more details on how we build the mappings and process the BC4GO entity linking dataset in Appendix~\ref{apx_dataset}.
We retrieve 30 candidates for each mention via querying the Gene Ontology API.
After the preprocessing, the expanded BC4GO dataset contains 29,037 mention-candidate pairs in the training set (9,027 positive and 20,010 negative pairs), 7,023 pairs in the dev set (2,352 positive and 4,671 negative pairs), and 5,580 pairs in the test set (1,578 positive and 4,002 negative pairs). During the evaluation, we set each mention with one candidate as one pair, and use accuracy to calculate the correct prediction pair number over the total number of mention-candidate pairs.

\begin{table}[t]
\begin{center}
\resizebox{0.85\columnwidth}{!}{%
\begin{tabular}{ l | c | c | c | c | c | c }
\toprule

GO number & 10 & 15 & 20 & 25 & \textbf{30} & 35  \\
\midrule
Fraction & 0.56  & 0.68 & 0.79 & 0.82 & 0.83 & 0.83 \\
\bottomrule
\end{tabular}
}
\end{center}
\caption{The fractions of mentions that can be found with at least one positive
candidate.}
\label{tab:fraction_candidate}
\end{table}








\subsection{Baselines}

\paragraph{Event Extraction}
We compare \jointEE{} with several recent state-of-the-art methods on biomedical event extraction, including: TEES~\cite{bjorne2011tees}, EventMine~\cite{Pyysalo2012eventmine}, Stacked generalization~\cite{majumder2016stackedGeneralization}, TEES-CNN~\cite{bjorne2018biomedical}, KB-driven Tree-LSTM~\cite{LSTM2019}, QA with BERT~\cite{QandA2020}, GEANet~\cite{graph2020biomedical}, BEESL~\cite{sequencelabel}, DeepEventMine~\cite{deepeventmine}, HANN~\cite{Zhao2021hann}, and CPJE~\cite{wang2022cpje}.

\paragraph{Entity Linking}

We compare our \jointEL{} with the following baselines: LATTE~\cite{zhu2020latte}, Bootleg~\cite{2020bootleg}, Fast Dual Encoder~\cite{dualencoder2021fast}, and PromptEL~\cite{zhu2022}. Note that some of the existing entity linking approaches~\cite{yuan2022generative} require to be explicitly grounded on other knowledge bases. They are not comparable with our approach and thus we did not include them in our experiments.



\subsection{Implementation Details}
We first train the base event extraction and entity linking models on Genia 2011 and BC4GO datasets, respectively. For the base event extraction and entity linking models, we use AdamW~\cite{loshchilov2018decoupled} optimizer with a learning rate of 5e-5, and a linear learning rate warm-up over the first 10\% of training steps is applied. The model is trained for 30 epochs with a batch size of 16. We stop the training of these two base models if they do not show a better performance for 5 consecutive epochs. 
For joint training (both \jointEE{} and \jointEL{}), we use learning rates 2e-5, 1e-5, 1e-5, 5e-6, 5e-6, 5e-6 for 6 rounds respectively with a batch size of 16. We stop the training of these two \jointEE{} and \jointEL{} models if they do not show a better performance for 5 consecutive epochs.

\section{Results and Discussions}

\subsection{Main Results}
\paragraph{Event extraction}
Table~\ref{tab:main_ee_results} shows the results of our approach \jointEE{} and the baselines on Genia 2011 dataset. \jointEE{} achieves significant improvement over all the strong baselines. Particularly, it outperforms the base model by 3.12\% F1 score and the previous state-of-the-art by 1.6\% F1 score, respectively. These results demonstrate the effectiveness of our joint learning framework and our approach effectively improves the base model that is not enhanced by the entity linking model.

\begin{table}[t]
\scriptsize

\resizebox{0.5\textwidth}{!}{%
\begin{tabular}{l  r  r  r}
\toprule
\textbf{Method} & \textbf{Precision} &
\textbf{Recall} &
\textbf{F1 Score} \\
\midrule
TEES~\cite{bjorne2011tees} & 57.65 & 49.56 & 53.30 \\
EventMine~\cite{Pyysalo2012eventmine} & 63.48 & 53.35 & 57.98\\
Stacked generalization~\cite{majumder2016stackedGeneralization}&66.46 &48.96 &56.38 \\
TEES-CNN~\cite{bjorne2018biomedical}  & 69.45 & 49.94 & 58.10\\
KB-driven Tree-LSTM~\cite{LSTM2019}  & 67.01 & 52.14 & 58.65\\ 
QA with BERT~\cite{QandA2020} & 59.33 & 57.37 & 58.33 \\
GEANet~\cite{graph2020biomedical}  & 64.61 & 56.11 & 60.06 \\
BEESL~\cite{sequencelabel}  & 69.72 & 53.00 & 60.22 \\
DeepEventMine~\cite{deepeventmine}  & 70.52 & 56.52 & 62.75 \\
HANN~\cite{Zhao2021hann} & 71.73 & 53.21 & 61.10\\
CPJE~\cite{wang2022cpje} & \textbf{72.62} & 53.33 & 61.50 \\
\midrule
Base-EE (Ours) & 68.20 & 55.73 & 61.23\\
\jointEE{} (Ours) & 69.66 & \textbf{59.75} & \textbf{64.35} \\

\bottomrule
\end{tabular}
}
\caption{Performance comparison of various event extraction approaches on the \textit{development} set of BioNLP Genia 2011. (\%). Bold highlights the highest performance among all the approaches.}
\label{tab:main_ee_results}
\end{table}

\paragraph{Entity linking}
Table~\ref{tab:main_el_results} shows the performance of various approaches for biomedical entity linking based on the test set of BC4GO. We observe \jointEL{} significantly outperforms the four strong baselines by more than 2.73\%. While our base entity linking model shares a similar architecture as~\cite{sun2021biomedical}, by incorporating the additional event features from the local context, the accuracy of \jointEL{} is improved by a large margin (3.73\% in accuracy), demonstrating the benefit of event-based features to entity linking.


\begin{table}[t]
\resizebox{1.0\columnwidth}{!}{%
\begin{tabular}{l r }
\toprule
 \textbf{Method} & \textbf{Accuracy (\%)} \\
\midrule
 LATTE~\cite{zhu2020latte} & 82.71\\
 Bootleg~\cite{2020bootleg} & 78.51\\
 Fast Dual Encoder~\cite{dualencoder2021fast} & 82.03\\
 PromptEL~\cite{zhu2022} & 81.32\\
\midrule
Base-EL (Ours) & 81.35\\
\jointEL{} (Ours) & \textbf{85.08} \\
\bottomrule
\end{tabular}
}    

\caption{Performance comparison of various entity linking approaches on BC4GO in terms of accuracy (\%). The best performance is highlighted in bold.}
\label{tab:main_el_results}
\end{table}
\begin{table}[t]
\begin{center}
\resizebox{0.75\columnwidth}{!}{%
\begin{tabular}{ c |  c |   c  }
\toprule
&\multicolumn{1}{c|}{Event Extraction} &\multicolumn{1}{c}{Entity Linking} \\
\midrule
Rounds & F1 Score & Accuracy \\
\midrule
Base  & 61.23  & 81.35\\
\midrule
1st & 62.48 & 83.38 \\
2nd & 63.85 & 84.06 \\
3rd & \textbf{64.35} & \textbf{85.08} \\
4th & 64.28 & \textbf{85.08} \\
5th & 64.15 & \textbf{85.08} \\
6th & 64.15 & \textbf{85.08} \\
\bottomrule
\end{tabular}
}
\end{center}
\caption{F1 score (\%) of event extraction on the Genia 2011 development set and the accuracy (\%) of entity linking on the test set of BC4GO at each round of joint training. The best performance is highlighted in bold.}
\label{tab:numofpairs}
\end{table}

\begin{table*}[t]
\scriptsize
\begin{center}
\resizebox{0.86\textwidth}{!}{%
\begin{tabular}{c | c  | c  | c | c | c | c | c }
\toprule
\textbf{Event type} & \textbf{Base} & \textbf{Round 1} & \textbf{Round 2} & \textbf{Round 3} & \textbf{Round 4} & \textbf{Round 5} & \textbf{Round 6} \\
\midrule
Gene expression        & 78.15 & 79.08 & 80.23 & 81.03 & 81.12 & 81.00 & 81.00\\
Transcription          & 69.46 & 70.60 & 71.14 & 72.08 & 72.08 & 72.08 & 72.08\\
Protein catabolism     & 74.57 & 75.06 & 75.88 & 76.38 & 76.38 & 76.38 & 76.38\\
Phosphorylation        & 83.67 & 84.12 & 84.95 & 85.67 & 85.67 & 85.67 & 85.67\\
Localization           & 80.30 & 80.75 & 81.07 & 81.30 & 81.30 & 81.30 & 81.30\\
\textbf{Simple events} & 76.73 & 77.92 & 78.65 & 78.79 & \textbf{78.99} & 78.77 & 78.77\\
\midrule
Binding                & 52.19 & 55.26 & 56.73 & \textbf{58.36} & \textbf{58.36} & \textbf{58.36} & \textbf{58.36}\\
\midrule
Regulation              & 45.52 & 46.03 & 47.94 & 49.03 & 48.53 & 48.53 & 48.53\\
Positive regulation     & 49.52 & 50.82 & 52.02 & 52.95 & 52.52 & 52.41 & 52.41\\
Negative regulation     & 57.48 & 57.78 & 58.09 & 58.52 & 58.33 & 58.33 & 58.33\\
\textbf{Complex events} & 50.84 & 51.54 & 52.68 & \textbf{53.83} & 53.42 & 53.31 & 53.31\\
\bottomrule
\end{tabular}
}
\end{center}
\caption{Results of event extraction on the Genia 2011 development set for each fine-grained event type and three categories (simple events, binding events, and complex events) at each round in terms of F1 score (\%). The best performance is highlighted in bold.}
\label{tab:detailrounds}
\end{table*}

\subsection{Analysis of Generalizability}
We further conducted an experiment to evaluate the generalizability of our framework on two additional datasets that are in another domain (i.e., drug-disease association) and are supported by another knowledge base, i.e., the Unified Medical Language System (UMLS).
Specifically, for biomedical entity linking, we validate our model on NCBI Disease corpus~\cite{dougan2014ncbi} the disease mentions and their Concepts of Unique Identifiers (CUI) in UMLS from a collection of 793 PubMed abstracts. For biomedical event extraction, we adopt the pharmacovigilance (PHAEDRA) dataset~\cite{thompson2018annotation} that contains 4 types of structured drug-disease event information with 3 argument roles. 

We want to verify that the drug-disease event information on PHAEDRA can provide extra information to help the entity linking task on NCBI (see results in Table~\ref{tab:main2_el_results}). On the other hand, we also want to verify that the meaning of the disease mentions (disorder entities in PHAEDRA) from the UMLS knowledge base will help the drug-disease event extraction on PHAEDRA (see results in Table~\ref{tab:main2_ee_results}). 
From both Table~\ref{tab:main2_el_results} and Table~\ref{tab:main2_ee_results}, our approach achieves promising performance improvement compared with the base models and previous baselines on both entity linking and event extraction tasks. Particularly, on NCBI, we outperform the previous state-of-the-art biomedical entity linking model by 0.32\% accuracy, while we obtain 0.78\% improvement on biomedical event extraction. 
The experiment effectively demonstrates that our approach can be adopted to various domains.


\begin{table}[t]
\resizebox{1.0\columnwidth}{!}{%
\begin{tabular}{l r }
\toprule
 \textbf{Method} & \textbf{F1 Score (\%)} \\
\midrule
 NormCo~\cite{wright2019normco} & 87.80 \\
 SparkNLP~\cite{kocaman2021biomedical} & 89.13\\
 BioLinkBERT~\cite{yasunaga2022linkbert} & 88.76\\
 ConNER~\cite{jeong2022enhancing} & 89.20\\
 CompactBioBERT~\cite{rohanian2023effectiveness} & 88.76\\
 
\midrule
Base-EL (Ours) & 85.58\\
\jointEL{} (Ours) & \textbf{89.52} \\
\bottomrule
\end{tabular}
}    

\caption{Performance comparison of various entity linking approaches on the \textit{test} set of NCBI Disease. Bold highlights the highest performance among all the approaches.
}
\label{tab:main2_el_results}
\end{table}

\begin{table}[t]
\scriptsize

\resizebox{0.5\textwidth}{!}{%
\begin{tabular}{l  r  }
\toprule
\textbf{Method} & 
\textbf{F1 Score(\%)} \\
\midrule
EventMine~\cite{Pyysalo2012eventmine} & 61.60 \\
HYPHEN~\cite{thompson2018annotation} & 65.00 \\
\midrule
Base-EE (Ours) & 61.29\\
\jointEE{} (Ours) & \textbf{65.78} \\

\bottomrule
\end{tabular}
}
\caption{Performance comparison of various event extraction approaches on the \textit{test} set of PHAEDRA. Bold highlights the highest performance among all the approaches.
}
\label{tab:main2_ee_results}
\end{table}



\subsection{Impact of the Number of Training Rounds}
Tables~\ref{tab:numofpairs} show the performance of both event extraction and entity linking at each round of joint training based on the EM-style iterative algorithm. We observe that the performance of both models gradually increases with more rounds of joint training and both models achieve the highest performance after 3 rounds. Compared with the base models, both \jointEE{} and \jointEL{} achieve significant improvements with a large marge: 3.12\% absolute F1 score gain for event extraction and 3.73\% absolute accuracy gain for entity linking, demonstrating the effectiveness of our joint learning framework. Table~\ref{tab:detailrounds} shows the event extraction performance (i.e., F1 Score) on each fine-grained event type and three event type categories (including simple events, binding events and complex events) at each round during joint training. As we can see, with 3-4 rounds of joint training, \jointEE{} achieves up to 2.26\%, 6.17\%, and 2.99\% absolute F1 score gain on the simple, binding and complex events, indicating that binding events benefit the most from the entity knowledge from external knowledge bases. This is consistent with our observation as many entity descriptions in the knowledge base indicate the binding functions of the entities. We also observe that, with more rounds of joint training, the performance of \jointEE{} decreases more on complex events which contain multiple arguments and nested events, such as \textit{regulation}, \textit{positive regulation}, and \textit{negative regulation}. 



\subsection{Qualitative Analysis}
Table~\ref{tab:improved_EE_results} shows three examples for which the event predictions are improved and corrected within the first 3 rounds of joint training. Taking the first sentence as an example, before the first round of joint training, \jointEE{} mistakenly predicts a \textit{Phosphorylation} event triggered by ``phospho'' with ``STAT3'' as the \textit{Theme} argument due to the misinterpretation of the sentence. However, the entity knowledge retrieved from the Gene Ontology (GO) by \jointEL{} indicates that ``STAT3 is a regulation of tyrosine STAT protein and BMP-6 is a regulation of BMP signaling pathway'', while the word "regulation" from both GO definitions helps better disambiguate the context during the 1st round of joint training and finally calibrates the previous wrong event predictions to the \textit{Regulation} event triggered by ``changes'' with two arguments: ``STAT3'' as the \textit{Theme} and ``BMP-6 as the \textit{Cause} argument.  

Similarly, Table~\ref{tab:improved_EL_results} also shows three examples for which the entity linking results are improved and corrected within the first 3 rounds of joint training. Taking the first sentence as an example, before the 1st round of joint training, \jointEL{} mistakenly links the entity mention ``\textit{UNC-75}'' to the entity defined by ``\textit{Positive regulation of synaptic transmission}''. However, by incorporating the event knowledge with joint training, especially knowing that ``\textit{UNC-75}'' is the \textit{Theme} of a \textit{Binding} event, \jointEL{} correctly links ``\textit{UNC-75}'' to the target entity defined by ``\textit{single-stranded RNA binding}'' in the Gene Ontology.

\begin{table*}[th]
\scriptsize
\begin{center}
\resizebox{\textwidth}{!}{%
\begin{tabular}{c | l }
\toprule
\textbf{Rounds} \\ 
1st & \underline{\textbf{Text:}} \ We did not observe any significant \textbf{changes} in the level of \textbf{phospho}-\textit{STAT3} or phospho-p38 upon \textit{BMP-6} treatment of B cells. \vspace{0.8mm} \\ 
& \underline{\textbf{Previous Results: }} \ \textbf{Event type}: Phosphorylation; \ \textbf{Trigger}: phospho; \ \textbf{Theme}: STAT3. \vspace{0.8mm}\\ 
& \underline{\textbf{Entity Knowledge from \jointEL{}:}} \ \textbf{STAT3}: regulation of tyrosine STAT protein; \ \textbf{BMP-6}:regulation of BMP signaling pathway. \vspace{0.8mm}\\ 
& \underline{\textbf{New Results: }} \ \textbf{Event type}: Regulation; \ \textbf{Trigger}: changes; \ \textbf{Theme}: STAT3, \ \textbf{Cause}: BMP-6.\vspace{0.8mm}\\
\midrule

 
2nd & \underline{\textbf{Text:}} \ Costimulation through \textit{CD28} and/or CD2 did not \textbf{modulate}, the CD3-dependent \textbf{phosphorylation} of HS1.\vspace{0.8mm}\\
&  \underline{\textbf{Previous Results: }} \ \textbf{Event type}: Phosphorylation; \ \textbf{Trigger}: phosphorylation; \ \textbf{Theme}: CD28.\vspace{0.8mm}\\ 
& \underline{\textbf{Entity Knowledge from \jointEL{}:}} \ \textbf{CD28}: immune response; \ \textbf{CD2}: regulation of CD4, CD25 regulatory T cell differentiation.\vspace{0.8mm}\\
& \underline{\textbf{New Results: }} \ \textbf{Event type}: Regulation; \ \textbf{Trigger}: modulate; \ \textbf{Theme}: phosphorylation (event), \ \textbf{Cause}: CD28.
\vspace{0.8mm}\\
\midrule

3rd & \underline{\textbf{Text:}} \ When tested its ability to block calcineurin-dependent signaling in cells, the pivotal promoter element for \textit{interleukin-2} gene \textbf{induction}.\vspace{0.8mm}\\
&  \underline{\textbf{Previous Results: }} \ \textbf{Event type}: Regulation; \ \textbf{Trigger}: induction; \ \textbf{Theme}: interleukin-2.\vspace{0.8mm}\\
& \underline{\textbf{Entity Knowledge from \jointEL{}:}} \ \textbf{interleukin-2}: plastid gene expression.\vspace{0.8mm}\\
& \underline{\textbf{New Results: }} \ \textbf{Event type}: Gene expression; \ \textbf{Trigger}: induction; \textbf{Theme}: interleukin-2. \vspace{0.8mm}\\



\bottomrule
\end{tabular}
}
\end{center}
\caption{Example sentences and results for event extraction at each round sampled from the Genia 2011 development set. For each sentence, before each round of joint training, the event prediction is not correct while after incorporating the entity knowledge from \jointEL{}, the errors are corrected with joint training. The bold words in each text highlight the candidate event triggers while the italic words highlight the candidate arguments predicted by \jointEE{}.}
\label{tab:improved_EE_results}
\end{table*}

\begin{table*}[th]
\scriptsize
\begin{center}
\resizebox{0.82\textwidth}{!}{%
\begin{tabular}{c | l }
\toprule
\textbf{Rounds} \\ 
1st & \underline{\textbf{Text:}} \ To determine the elements in the exon 7 region that \textit{\textbf{UNC-75}} directly and specifically recognizes in vitro.\vspace{0.8mm}\\
& \underline{\textbf{Previous Results: }} \ \textbf{Entity:} Positive regulation of synaptic transmission.\vspace{0.8mm}\\
& \underline{\textbf{Event Knowledge from \jointEE{}:}} \ \textbf{Event type}: Binding; \ \textbf{Trigger}: recognize; \ \textbf{Theme}: UNC-75. \vspace{0.8mm}\\
& \underline{\textbf{New Results: }} \ \textbf{Entity:} single-stranded RNA binding.\vspace{0.8mm}\\
\midrule

2nd & \underline{\textbf{Text:}} \ These data suggest that ectopically expressed \textit{\textbf{BP}} downregulates FIL activity genes in ovaries.\vspace{0.8mm}\\
& \underline{\textbf{Previous Results: }} \ \textbf{Entity:} DNA-binding transcription factor activity, RNA polymerase II-specific.\vspace{0.8mm}\\ 
& \underline{\textbf{Event Knowledge from \jointEE{}:}} \ \textbf{Event type}: negative regulation; \ \textbf{Trigger}: downregulates; \ \textbf{Theme}: FIL.\vspace{0.8mm}\\
& \underline{\textbf{New Results: }} \ \textbf{Entity:} negative regulation of gene expression.
\vspace{0.8mm}
\\
\midrule

3rd & \underline{\textbf{Text:}} \ Effects of the \textit{\textbf{gei-8}} mutation on gene expression were studied with whole genome microarrays.\vspace{0.8mm}\\
& \underline{\textbf{Previous Results: }} \ \textbf{Entity:} reciprocal meiotic recombination.\vspace{0.8mm}\\
& \underline{\textbf{Event Knowledge from \jointEE{}:}} \ \textbf{Event type}: gene expression; \ \textbf{trigger}: expression; \ \textbf{Theme}: gei-8.\vspace{0.8mm}\\
& \underline{\textbf{New Results: }} \ \textbf{Entity:} transcription by RNA polymerase II.\vspace{0.8mm}\\
\bottomrule
\end{tabular}
}
\end{center}
\caption{Example sentences and results for entity linking at each round sampled from the development set of BC4GO dataset. For each sentence, before each round of joint training, the entity linking result is not correct while after incorporating the event knowledge from \jointEE{}, the errors are corrected with joint training. The bold words in each text highlight the candidate entity mention for entity linking.}
\label{tab:improved_EL_results}
\end{table*}

\subsection{Error Analysis}
 
We further sample 50 prediction errors for both entity linking and event extraction based on their results on the development set of each dataset, respectively. We summarize the main error categories for each task as follows: 




\paragraph{Event Extraction} The main error (33/50) for event extraction lies in the missing or spurious argument predictions. Most event types such as simple events (including \textit{gene expression}, \textit{transcription}, \textit{localization}, \textit{phosphorylation}, and \textit{protein catabolism}) are defined with a fixed number of arguments while the 
complex events and binding events are usually associated with up to four possible arguments, thus the model tends to miss some arguments or predict spurious arguments. Taking the following two sentences as examples:

\begin{itemize}

\item \textbf{S1:} The \textbf{FOXP3} (\textbf{\textit{arg: Theme}}) \textbf{inhibition} (\textbf{\textit{trigger: Negative regulation}}) by GATA element in the FOXP3 \underline{\textbf{promoter}} \underline{\textbf{(redundant arg: Site)}}.

\vspace{2mm}

\item \textbf{S2:} Disruption of the \textbf{Jak1} (\textbf{\textit{arg: Theme}}) \textbf{binding} (\textbf{\textit{trigger: Binding}}), proline-rich \textbf{Box1} (\textbf{\textit{arg: Site}}) region of \underline{\textbf{IL-4R}} \underline{\textbf{(missing arg: Theme)}} abolished signaling by this chimeric receptor.
\end{itemize}

For S1, our model successfully predicts \textit{inhibition} as a Negative regulation event and \textit{FOXP3} as its \textit{Theme} argument. However, it also mistakenly predicts \textit{promoter} as a \textit{Site} argument, due to two possible reasons: (1) the entity \textit{promoter} is frequently labeled as a \textit{Site} argument in the training set; and (2) the protein \textit{FOXP3} is defined as ``\textit{regulation of DNA-templated transcription}'' in the Gene Ontology, which also tends to imply \textit{promoter} as a \textit{Site} argument. In S2, our \jointEE{} correctly predicts \textit{binding} as a \textit{Binding} event with two arguments: \textit{Jak1} and \textit{Box1}. However, it mistakenly misses another \textit{Theme} argument which is likely because the model treats \textit{IL-4R} as \textit{Box1} which is already labeled as a \textit{Site} argument.

\paragraph{Entity Linking} 76\% (38/50) of the remaining error for entity linking lies in the candidate retrieval where the candidate sets retrieved based on the Gene Ontology (GO) API for some entity mentions do not contain their true target entities. For example, for the entity mention ``\textit{TAT-DeltaDBD-GATA3}'', the candidate set returned by GO API does not include the true target entity GO:0019799 with name \textit{acetyl-CoA:alpha-tubulin-L-lysine 6-N-acetyltransferase activity}. 

\section{Conclusion}


In this work, we propose a joint biomedical entity linking and event extraction framework, i.e., \jointEL and \jointEE, to leverage the benefit of one task to the other. Our \jointEE can incorporate the domain knowledge obtained by \jointEL, while \jointEL{} can be improved by the event structural context provided by \jointEE. To iteratively improve the two tasks together, we propose a novel hard-EM-style learning strategy where we first estimate missing variables for both two incomplete datasets based on the current task-specific models, and then update the parameters of both models on the datasets that are augmented by pseudo labels.
We conduct extensive experiments on the biomedical entity linking dataset, i.e., BC4GO, and biomedical event extraction, i.e., Genia 11. We also provide several valuable discussions such as error analysis that reveals the remaining challenges of both two tasks. We hope this work can shed light on
the following research on biomedical information extraction and broader communities.

\section*{Limitations}

While our proposed joint framework and learning strategy showcase promising results for entity linking and event extraction in the biomedical domain, one limitation is the potential restriction of specific domains and knowledge bases which both entity linking and event extraction tasks can share with. 

\bibliography{anthology,custom}
\bibliographystyle{acl_natbib}

\appendix
\section{More Details on Experiment Setup}

\subsection{Dataset Processing for Entity Linking}
\label{apx_dataset}
In this section, we elaborate on how we construct the mappings between the mentions in BC4GO and the entities in Gene Ontology.
Specifically, 
Gene Ontology consists of three directed acyclic graphs (DAG) and there is no connected component between them.
All the nodes in the three DAGs are biomedical concepts and each DAG has a hierarchical structure called an \textit{ancestor chart}. The biomedical concepts at the upper level of an \textit{ancestor chart} have broader meanings compared with the biomedical concepts at the lower level. The three nodes at the top of the three \textit{ancestor chart}s in Gene Ontology are: \textit{Cellular Component}, \textit{Molecular Function}, and \textit{Biological Process}, which are the broadest concepts in biomedical science and any biomedical concept is contained in one of them. We refer to the three topmost nodes in the three \textit{ancestor chart}s as root nodes in the rest of the section.  
For each node in an \textit{ancestor chart}, the nodes above it are broader concepts while the nodes below it are finer concepts, and there is no edge between the nodes from the same level. 
For each entity mention $m_i$ from a sentence \vvv{W} of the BC4GO dataset, we denote the annotated target entity from Gene Ontology for $m_i$ in the original BC4GO dataset as $c^{gold}_{i}$. We also take $m_i$ as a query to the search engine via the API of Gene Ontology and obtain a set of candidate entities $\mathcal{C}_i=\{c_{i1},c_{i2},...,c_{ij}\}$. Then, for each $c_{ij}\in\mathcal{C}_i$, we find a path between $c_{ij}$ and the root node of a particular \textit{ancestor chart}, denoted as $\vvv{P}_{ij}$. Following a similar process, we also find the path $\vvv{P}_{gold}$ between the gold target entity $c^{gold}_{i}$ and a root node. Based on these paths, we design the following four strategies to determine a new set of gold entities from $\mathcal{C}_i$ for each mention $m_i$ as the reference in Gene Ontology:
\begin{itemize}
  \item If a candidate entity $c_{ij}\in\mathcal{C}_i$ for the mention $m_i$ is the same as the ground-truth concept $c^{gold}_{i}$ in the original BC4GO dataset, we add $c_{ij}$ to the gold target entity set.
  \item If a candidate entity $c_{ij}\in\mathcal{C}_i$ for the mention $m_i$ is on the path $\vvv{P}_{gold}$, we add $c_{ij}$ to the gold target entity set.
  \item If the gold target entity $c^{gold}_{i}$ annotated in the original BC4GO dataset is on the path $\vvv{P}_{ij}$ for a candidate entity $c_{ij}$, we add $c_{ij}$ to the gold target entity set.
  \item If the path $\vvv{P}_{ij}$ for a particular candidate entity $c_{ij}$ have more than 4 overlapped nodes with the path $\vvv{P}_{gold}$ for the original gold entity, we add $c_{ij}$ to the gold target entity set.  
\end{itemize}

\end{document}